\documentclass[conference]{IEEEtran}
\usepackage{blindtext, graphicx,amsmath}
\usepackage[caption=false]{subfig}
\usepackage{amsmath}
\usepackage{xcolor}
\usepackage[linesnumbered,ruled,vlined]{algorithm2e}

\SetCommentSty{mycommfont}

\graphicspath{{Images/}}

\ifCLASSINFOpdf
\else
\fi

\begin{document}

\title{ Efficient resource management in UAVs \\for Visual Assistance}

\author{\IEEEauthorblockN{Bapi Reddy Karri}

 }

\maketitle

\begin{abstract}
There is an increased interest in the use of Unmanned Aerial Vehicles (UAVs) for agriculture, military, disaster management and aerial photography around the world. UAVs are scalable, flexible and are useful in various environments where direct human intervention is difficult. In general, the use of UAVs with cameras  mounted to them has increased in number due to their wide range of applications in real life scenarios. With the advent of deep learning models in computer vision many models have shown great success in visual tasks. But most of evaluation models are done on high end CPUs and GPUs. One of major challenges in using UAVs for Visual Assistance tasks in real time is managing the memory usage and power consumption of the these tasks which are computationally intensive and are difficult to be performed on low end processor board of the UAV. This projects describes a novel method to optimize the general image processing tasks like object tracking and object detection for UAV hardware in real time scenarios  without affecting the flight time and  not tampering the latency and accuracy of these models. 
\end{abstract}


\IEEEpeerreviewmaketitle

\section{Introduction}
The UAVs stand for Unmanned Aerial Vehicles, also known drones are air-crafts with no human on board. In general UAVs are remote controlled aircraft or can fly autonomously based on pre-programmed flight plans or more complex dynamic automation systems\cite{7347786,7838688}. The UAVs can be easily integrated with sensors to accommodate specific needs in a short span of time. This gain in popularity of UAVs can be attributed to the miniaturization of electronics and easy availability of portable and low-power sensor solutions\cite{7824758}. This lead to application of UAVs in various fields like agriculture, robotics, filming and military related activities. Among all these fields UAVs are mounted with cameras to analyze their surrounding environments. In order to analyze the images captured by these cameras the UAV should have a real time image processing model.

\par Generally when a drone is flown in an area of interest to analyze a particular environment, the drone is assigned some task which involves understanding of the other objects which are present in the environment. Let us consider a disaster hit where the UAV needs to deploy communication nodes. In this scenario the UAV needs to analyze whether there are people present in area of deployment which means that drone needs to detect the region where maximum people are present and deploy the node there. This task can be achieved by performing object detection in current scenario. Now lets us scale this idea instead of using one drone for deployment we would like to use multiple drones so that we can deploy the nodes in affected area quickly. Here if we consider a swarm model where one drone co-ordinates other drones as shown in \ref{fig:swarm}. And the slave drones need to adjust their position by tracking their master drone. This task can be framed as a object tracking problem where the slave drone needs to track the master drone's position time to time. 
\section{Object Tracking}
\begin{figure}[h]
    \centering
    \includegraphics[scale=0.4]{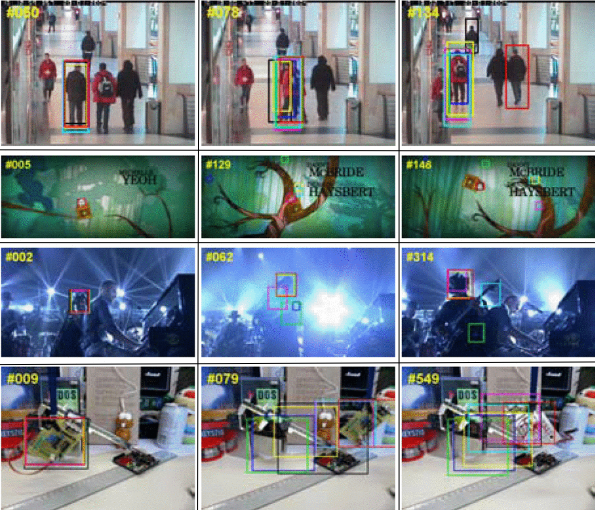}
    \caption{Tracking in dynamic environments by various trackers}
    \label{fig:dynamic-environment}
\end{figure}

Object tracking is one of the fundamental problems of computer vision. Object tracking in simple terms means locating the object over a series of image frames. Even though object tracking may sound simply tracking object it's a broad term which encompasses similar in concept but technically different ideas. Even with many algorithms available for object tracking depending upon the use case the challenges may vary like change of illumination, pose variation of the object and heavy occlusion \cite{8049446}. As shown in \ref{fig:dynamic-environment} we can see that as the dynamics of scene changes the tracked regions are changed.
There are various methods to solve these problem like optical flow algorithms which track the location of featured pixels in image\cite{6392450}, Kalman Filtering which is predication based algorithm for locating moving objects based on motion information extracted form series of frames \cite{7056449}. Next are Meanshift and Camshift algorithms which are used for locating the maximum shift for predicting the moving object. And then there are single object trackers where a area is marked and the object is tracked using any of the tracking  algorithms like BOOSTING, MIL, KCF, TLD, MEDIAN FLOW.

\par
\begin{figure}[b]
    \centering
    \includegraphics[width=\linewidth]{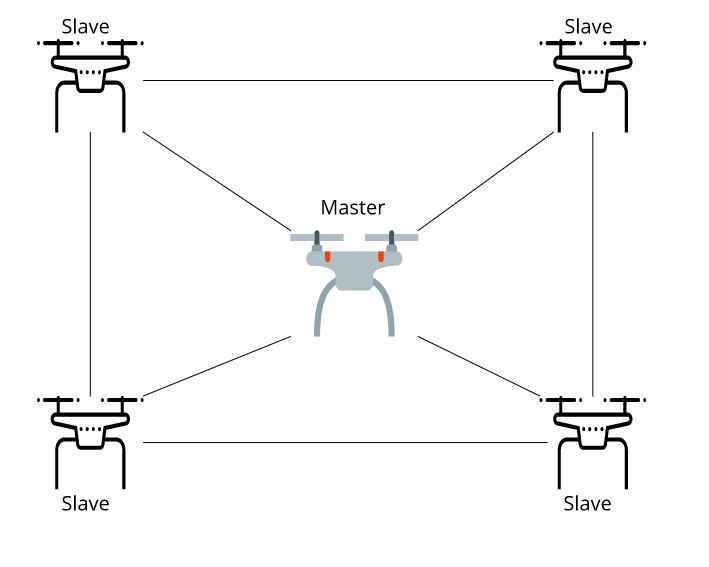}
    \caption{swarm of drones}
    \label{fig:swarm}
\end{figure}

 In recent years there are new trackers based on learning algorithms with improved accuracy. They are more accurate than the general KCF,TLD trackers. Like a SCM tracker which uses a sparsity based discriminative classifier on sparsity based general model \cite{6247882}. Also, there are trackers for tracking the motion of object based on deep learning like DLT tracker which uses features extracted from deep learning architectures to calculate the trajectory of object \cite{NIPS2013_5192}. In case of UAVs, about object tracking can be used in various scenarios like autonomous surveillance, swarm co-ordination. Let us consider swarm co-ordination scenario as shown in \ref{fig:swarm}.
Here the slave UAVs needs to keep a track of master UAV. Instead of calculating the whole region in which the object is present it is better to know the direction in which the master drone is moving. Now one of the issues in using the trackers is that they give us the area i.e., region instead of the direction in which the object is moving. For slave UAVs when we use a object tracker we would like to know the direction in which the object is flying and move according to that direction. So in order to know the direction of the moving object from the tracker we need to have a direction prediction scheme. We can calculate the optical flow of the object using the tracker to predict the direction. But they are not suitable in our scenario as the number of operations performed is very high. Also, the latency produced in the system will also be very high when we take a video input and process as frame by frame basis. In order to overcome these problems in UAVs which have low end processors boards we propose a direction prediction scheme based on the BFS algorithm on the roi outputs of various trackers. This method is explained further in the \textbf{section 6.1}.


\section{Object Detection}

\begin{figure}[h]
    \centering
    \includegraphics[width=150pt]{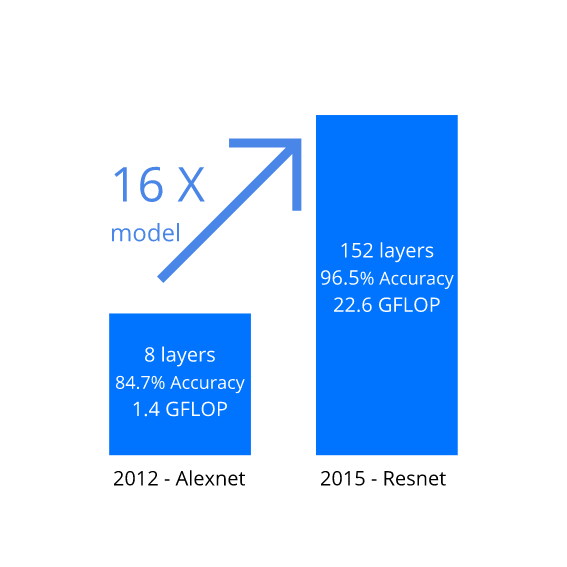}
    \caption{Image Recognition Models comparison}
    \label{fig:models-comparision}
\end{figure}

Object detection is one of the core problems in computer vision that people are trying to solve from a very long time. Current state of art object detectors are based on deep convolutional neural networks. The performance of these deep neural networks has increased from 84.7\% in 2012 by AlexNet \cite{NIPS2012_4824} to 96.5\% in 2015 by ResNet \cite{DBLP:journals/corr/HeZRS15} for image recoginition tasks. Such high accuracy enables these models to bring artificial intelligence to far-reaching applications not only in UAV's but also in self driving cars. But high accuracy at these levels comes at the cost of high computational complexity. From \ref{fig:models-comparision} we can see that the size of models has increased the number of operations drastically. As these models are developed considering accuracy as high priority compared to latency their deployment in real time embedded devices like drones causes a huge hurdle. In UAVs we have limited computational resources and in most cases are battery constrained it is difficult to deploy these models and achieve these real time performances.
\par Before going into the specifics of the computational complexity of these models let us first the understand the approaches of object detection problem in general. In general two approaches are followed for object detection one is sliding window approach where a sliding window is selected from list of region proposals of a image and it is passed through deep CNN to find the category of the object. Now one issue with this approach is that the region proposals are very large in number around \textbf{2000} in general. So passing each crop from the region proposal leads to a heavy latency in detection. R-CNN based detectors belong to this category.

Now the other category of detectors work by dividing the image into a coarse \textbf{NxN} grid where each box has base bounding boxes. For each bounding box we predict offset of true location from this bounding box and also classification scores for each bounding box. And then apply threshold by removing bounding with low classification score and low offset score.Single Shot Multibox Detector and Yolo detector belong to this class of detectors.
\par Although these models have high accuracy and fps these results are obtained on high-end GPU's and CPU's. There are two main problems in deploying these DNN models directly. 
\begin{enumerate}
\item Power constraint: 
UAVs are power constrained in general with limited battery available for flight time so the heavy computations performed by DNN models will quickly drain the battery.
\begin{equation}
    f_o * N_os * E_o = P_w
    \label{eqn:power}
\end{equation}
From \ref{eqn:power} if we assume that processor board operates at 10Hz and number of operations performed are generally in magnitude of billions so lets take 1 billion operations per sec and energy consumed for each operation is around 600pJ then power consumed is around \textbf{3W}, just for storing the model. This rate of power consumption is beyond the scope of normal drones which operated with batteries in the range of 21W-40W. Even if we simply try to test these models it will lead to lesser flight times. 
\item Memory constraint:
The processor board which we are using the UAVs has a RAM around 1GB. The typical size of these DNN models is around 100MB-300MB. And when we are running these models in real time they also need to store the images in order to process them in real time. Now when are performing object detection we need to load these models into RAM and pass the input image from camera feed to these model continuously which is a computationally intensive task with high latency. As \textbf{40\%} of available RAM is just being used only to store the model the scope to perform other tasks on UAV is very little which means that the UAV cannot take any control decisions during this period. Therefore to perform detection the UAV must remain in stable position before loading the model. 
\end{enumerate}
So in order to support these models on the low end embedded processor we propose a scheme in \textbf{section 6.2} which deals with compression of these large DNN models preserving their accuracy. And also a scheduling scheme taking advantage of the multi cores present in the low end processor board of UAV. The main idea here is to decrease the number of operations performed by the model to detect the object so that we can preserve power to continue the flight the larger distances.
\section{Related Works}

This section consists of some relevant works from which our work draws inspiration. There are mainly two category of works one where DNN models are used in UAVs for various scenarios. In the other customized hardware is built for efficient implementation of the DNN models in real time systems. 

\subsection{Deep learning model Applications}
UAV play a crucial role in aerial surveying, photography and payload dropping. When UAV are deployed in certain scenarios they need to analyze their environment and act according to the constraints present. So in order to take decisions when performing the UAV should have some understanding of their surrounding environments which is provided by sensors attached to the UAVs. This data produced by sensors is use by deep neural networks to make such decisions. With recent development in deep learning it became possible to solve some crucial challenges in various fields using UAVs as agents. Like, in package-delivery, dropping a payload in different climatic conditions has been achieved \cite{8023651}. In area of marine biology, we can now study the life cycles of different animals by tracking with drones \cite{7358813}. Also it is now possible to do aerial surveying with more accuracy in identifying people \cite{7515608}. Now one of the common theme among these researches is that they focus on applying the deep neural networks in real scenarios using UAVs. The focus is more on neural network part compared to that of constrains in the hardware components like power consumption and flight time of UAVs. Here the algorithm are optimized for custom challenges which they are facing, but in order to run these algorithms you need a efficient frame work to run the computational tasks of DNN's.

\subsection{Custom Hardware Architectures}
There was a rapid progress in custom hardware architecture for deep learning models. Researchers have designed custom hardware architectures for specialized neural networks \cite{5981829,7753296,7995253}. With these architectures they were able to achieve higher efficiency compared to that of CPU's and GPU's. Initial set of architectures implemented basic architectural components similar to Harvard Architecture. Next they realized that accessing memory is the crucial challenge, so the next set of accelerators efficiently optimized steps involving memory transfer and data movement \cite{7998996,7011421,7738524}. These initial movement of new set of architectures helped in increasing the efficiency and speed of DNN's. These specialized architectures were implemented using FPGA which were more robust than general GPU's used for DNNs. However even with these developments these architecture have not made it to UAVs due to focus solely on the DNNs and are not suitable for multipurpose tasks in UAVs.
\par Even with recent progress in implementation of DNNs on UAVs and rapid progress in Hardware architectures they are not many hardware architectures custom built for UAVs and dnns to meet the real time constraints. As the first set of methods consider UAVs only as a mode of application and second set of methods consider DNNs as only tasks which are being operated on the hardware. Both of these methods treat the either DNN algorithms as black-box or the control part of UAV as black box. However, in our research we have found that we can customize the DNN models to low end processor boards and meet the real time constraints of flight time and latency of the models.

\section{Our Method}
\begin{figure}[h]
    \centering
    \includegraphics[width=0.85\linewidth]{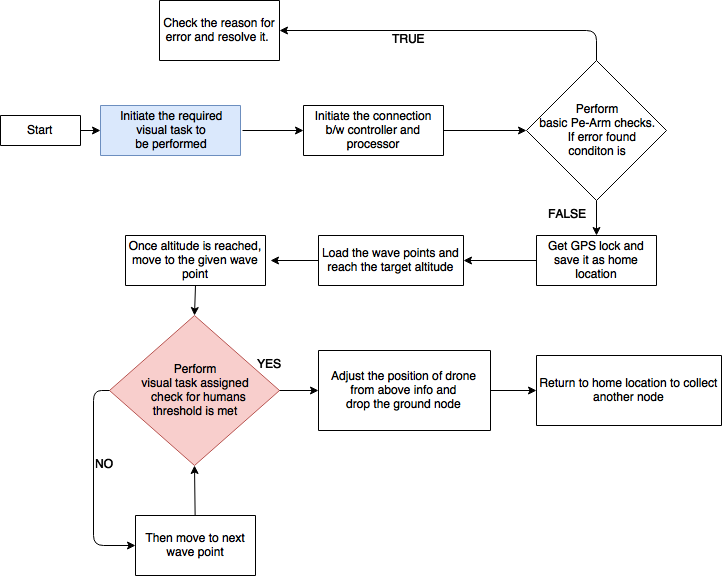}
    \caption{Task sequence for Node deployment.}
    \label{fig:flightPath}
\end{figure}

In this project our aim is to use the computing resources of the drone as efficiently as possible for visual tasks. \ref{fig:flightPath} shows the general sequence of tasks performed on UAVs which we are following for node deployment task.
In the visual task we perform either object tracking or object detection depending upon the scenario. Now we describe two methods where we decrease the number of times a operation is being performed for object tracking and another method to decrease the number of operations for object detection model. 
\subsection{Reduction in number of cycles}
As we have previously discussed the importance of object tracking and object detection methods we will see how we can improve their efficiency in UAVs. In object tracking we have to estimate location of visual target in each frame of an image sequence. Now when we are using object trackers in drones we would like to follow the object movement i.e., instead of finding the location of moving object we would like to know its direction of movement. The current state of object trackers initially start with a region of interest and find the correlation b/w bounding boxes using a sliding window approach. Now, in order to predict the location of moving object in the next frame the numbers of operations are very huge depending upon the size of the image. Instead of that we frame direction prediction as a regression problem where we need to find the co-ordinate of the center of moving sub-image. In a single pass we can find the most probable direction in which the object is moving.
\begin{algorithm}
    \SetKwInOut{Input}{Input}
    \SetKwInOut{Output}{Output}
    \SetKwInput{Initialization}{Initialisation}
    \SetKwProg{Fn}{}{}{}\SetKwFunction{DirectionPredictor}{Direction-Predictor}%
    \SetKwProg{Fn}{}{}{}\SetKwFunction{BFS}{Breadth-First-Search}%
\caption{Direction Predictor}
    \Input{$X_0,Y_0$ -Co-ordinates of background ROI,\\
           $Frame$ - 2D Matrix representing pixel values \\of image.}
    \Output{$Direction = {Top, Bottom, Left, Right}$}
    \Fn {\DirectionPredictor {$X,Y,Frame$}}{
        \Fn {\BFS{$X_0,Y_0,Frame$}} {
         \tcp{Set of adjacent frame centres at layer:$r$}
        $F^r = [f_1^r,f_2^r,f_3^r,...f_n^r]$ \\
        \tcp{List of all layers from point $x,y$ }
        $L_{x,y} = [F_{x,y}^0,F_{x,y}^1,F_{x,y}^2,F_{x,y}^3...F_{x,y}^k]$\\
        \Initialization{$Queue Q = \emptyset, r = 0 $}
         Q.push($L_{x,y}[r]$)\\
        \While{$Q \neq \emptyset$}{ 
         TempList = Q.pop()\\
         \tcp{Max Correlation function gives the frame center with maximum correlation.}
         p,q =   Max-Correltion(TempList);\\ 
         \eIf{Correlation(p,q) $>$ Threshold}{
            \textbf{return (p,q)};
         }
         {r = r+1;\\
         Q.push($L_{p,q}[r]$);
         }
        }
       }
       $ \delta x = p-X_0,\delta y = q-Y_0;$\\
       \tcp{Update the background center coordinates}
       $X_0 = \alpha.p + (1-\alpha).X_0$\\
       $Y_0 = \beta.q + (1-\beta).Y_0$ \\
       return direction based on \underline{$\delta x$} and \underline{$\delta y;$}
     }
    \label{algo:direct-predict} 
\end{algorithm}
With the algorithm \ref{algo:direct-predict} the number of operations performed in order to predict the direction are far less compared to that of general object trackers as here we are first assigning probabilities to the nearest neighbours of the previous co-ordinates and then moving on to next layers only when probability due to this layer I is less compared to previous probability. Also in most scenarios the object will be having a smooth transition rather than abrupt or sudden change in location so the maximum number of layers moved by image are far less compared to number of windows in sliding window approach.

\subsection{Reduction in number of operations}
\begin{figure}[htp]
\centering
 \subfloat[Input depth wise convolutional filters]{
	\label{subfig:depthconvo}
	\includegraphics[width=0.75\linewidth]{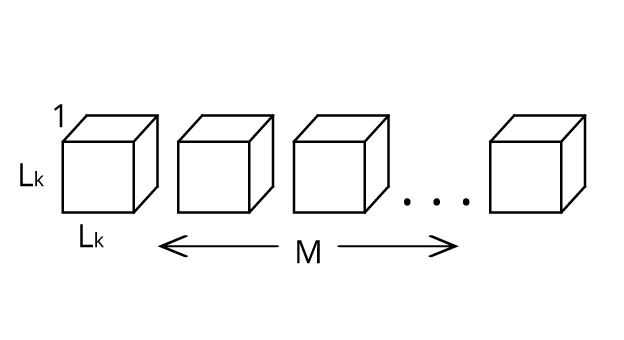} 
	} 
\hfill
\subfloat[1×1 Convolutional Filters called Point-wise convolution for depth wise convolution.]{
	\label{subfig:pointwise convo}
	\includegraphics[width=0.75\linewidth]{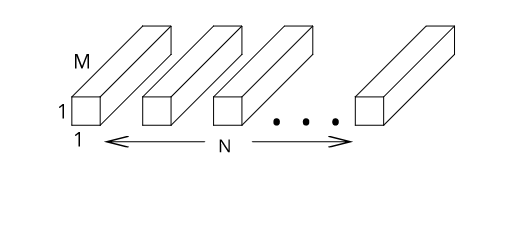} 
	}
\centering	
\caption{Two step Process}
\label{fig:twostep}
\end{figure}

Now when we can decrease the number of steps in which the model operates we can use the algorithm mentioned in above subsection. What if, we number of steps the operation is performed is only once, but number of operations is very large i.e., a billion operations for single frame of image. This leads us to the tasks of object detection which involve DNNs with multiple layers and pipeline networks. Consider a scenario where you want to study the objects present in a environment, now with a camera mounted drone you can take the image and find the objects of your interest from the image. But, if you want to do some real time analysis then we need to implement the object detection in real time in drone instead of just capturing the image and processing them later. This is where object detectors help us to solve the problem, initially we train a detector with the classes of our required object  and then we use this classifier in real time for detection. But one of the issues with these classifiers is that they are large in size i.e., the number of weights and operations to be performed are very huge in number. So in this even though the number of cycles of the operation performed is once the operation which is being performed is computationally intensive task. In order to do that we need to understand some basic concepts of Convolution neural networks architecture which are used for object recognition tasks. A convolutional neural network consists of filters which are used to extract features for recognizing objects in image. For images we will use spatial filters which are two dimensional in nature. Now in general object detection we have these deep convolutional networks as building blocks as shown in below figure:\\
\begin{figure}[h]
\centering
\includegraphics[width=\linewidth]{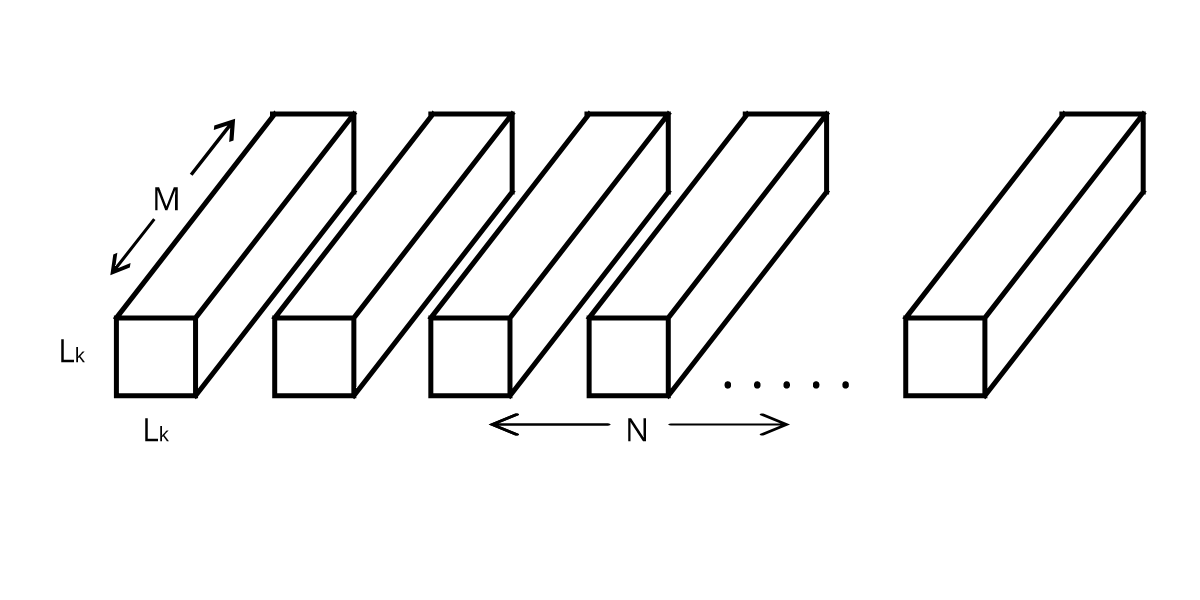}
\caption{General convolutional Neural Network}
\label{fig:standardconvo}
\end{figure}
\\Here M = "Size of input channel", N = "Size of output of channel", and \( L_k*L_k \) = "Size of Kernel".\\Let the input feature map be \textbf{I} with size of \(L_f*L_f*M \) and output feature map produced by the above convolutional network be \textbf{O} with size of \( L_f*L_f*N \) .Now the Kernel which is being used for this convolution be \textbf{K} is of size \( L_k*L_k*M*N \). Now the input and output feature maps are related as:
\begin{equation}
O_{p,q,n}= \sum_{i,j,m} K_{i,j,m,n} . I_{p+i-1,q+j-1,m,n}  
\end{equation}
Now the number of operations performed for calculating the output are :
\begin{equation} 
L_k * L_k * M * N * L_f * L_f 
\end{equation}
In general convolutional neural network generate the features based on kernels and combines these features to produce the output. Instead of having two operations in same step we can divide this into two independent steps as shown in Figure ~\ref{fig:twostep}
With the help of factorized convolution we can split the filtering and combining steps into depth-wise separable convolution with reduction in number of operations performed. Depth-wise separable convolutions consists of two layers: depth-wise convolution and \(1x1\) point-wise convolution. This idea of dividing the standard convolution steps into two separate steps is inspired from mobilenet architecture \cite{DBLP:journals/corr/HeZRS15}. In first layer we apply a separate filter for each input channel traversing the whole input depth.  convolution. At this stage input and output feature maps are related as:
\[ \hat{O}_{p,q,m} = \sum_{i,j,m} \hat{K}_{i,j,m} . I_{p+i-1,q+j-1,m}  \]
where \textbf{$\hat{O}$} is the output feature map and \textbf{$\hat{K}$} is the depth-wise convolutional kernel with size \( L_f*L_f*M \). Number of operations performed at this step is:
\begin{equation}
 L_k . L_k . M . L_f . L_f
 \end{equation}

Now with these features as inputs a linear combination is applied i.e., \(1*1\)
point-wise convolution. Here in linear combination each kernel value is multiplied with corresponding weight. Therefore number of operations at this step:
\begin{equation}
M . L_f . L_f . N 
\end{equation}
With these separation of steps the reduction in number of steps obtained is:
\begin{equation}
\begin{split}
(L_k.L_k.M.N.L_f.L_f)/(L_k.L_k.M.L_f.L_f +M.L_f.L_f.N)  \\ = 1/N + 1/L_k^2 
\end{split}
\end{equation}
Now that we were able to reduce the number of operations, we later observed that preserving same level of accuracy is not possible.

\section{Implementation}
Mobile-net architectures achieves similar results to that of standard convolution models while using only minimum of 50\% less parameters. When the size of models are compared, mobilenet architecture are order of \textbf{10 times}  smaller compared to standard convolution models. This is really helps in case of UAVs. In terms of hardware it means that we can have great speed and energy efficiency compared to that of standard models, as the number of operations performed are less. Firstly, speed or low-latency of the system achieved by mobilenet model is crucial in UAVs as in real time we need to take a real time decisions. From computational side, less model size means the number of operations that need to performed are far less so latency will be low enough. Second reason when we have large models it requires more memory to fetch these weights of the model which is very crucial as drop in power level causes change in position of UAVs. And also as UAVs are battery constrained it is difficult to run these models for long time. For CPU, the number of fetches required to load the mobilenet model are far less compared to standard convolution models.\par 
\begin{figure}[h]
\centering
	\label{fig:generalflow}
	\includegraphics[width=0.5\linewidth]{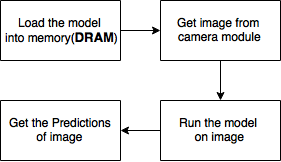} 
	\caption{General computational model for DNNs.}
\end{figure} 
We introduce a new scheduling model to take advantage of the small size of the mobilenet model as it can be loaded into SRAM quickly compared to standard convolution models. General DNN models are implemented as show in Fig \ref{fig:generalflow}.
\par
\begin{figure}[h]
\centering
	\label{fig:parallelflow}
	\includegraphics[width=0.5\linewidth]{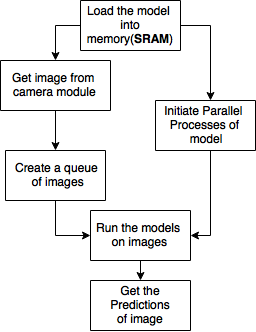}
	\caption{Parallel Processes computational model for DNNs.}
\end{figure}

 In case of mobilenet as the size of models is small we can store most of the model in SRAM of the CPU, so that it can be accessed easily. Now the processor boards used in the UAVs are generally multi-core processors. In Fig\ref{fig:parallelflow} we propose a new scheduling scheme for performing the computation of our mobilenet convolution models i.e, we can take advantage of the small size model by loading it into the memory and then initializing parallel tasks on each core such that they whenever an image is given from we pass it into the queue.
In the mean time the CPU creates parallel tasks one on each of its cores. When we queue is filled each image from the top of queue is passed to the parallel task on each core where they get processed, later the predictions are annotated on the respective images and sent to output queue for display.

\section{Results}
We have implemented the updated tracker algorithm for SCM, DLT, Struck and HCF trackers:
    \begin{figure}[h]
\centering
	\label{fig:tracker}
	\includegraphics[width=\linewidth]{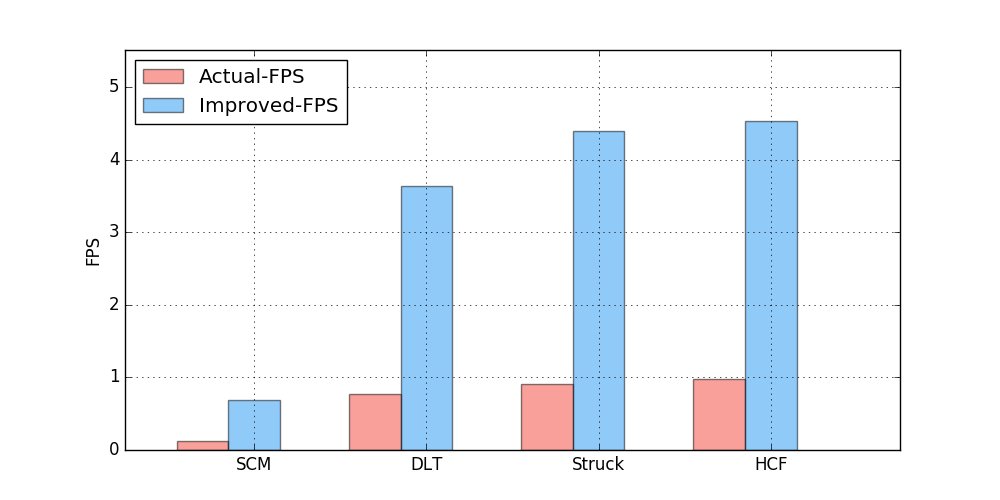} 
	\caption{FPS comparison b/w trackers with our method}
\end{figure} 
As the above trackers are slow in real time systems we have selected them to see the improvement in their FPS after application of our system. From our implementations we observed that there is an improvement around \textbf{4-6 times} for the trackers depending upon their tracking algorithm.
\begin{figure}[h]
\centering
	\label{fig:TPR}
	\includegraphics[width=\linewidth]{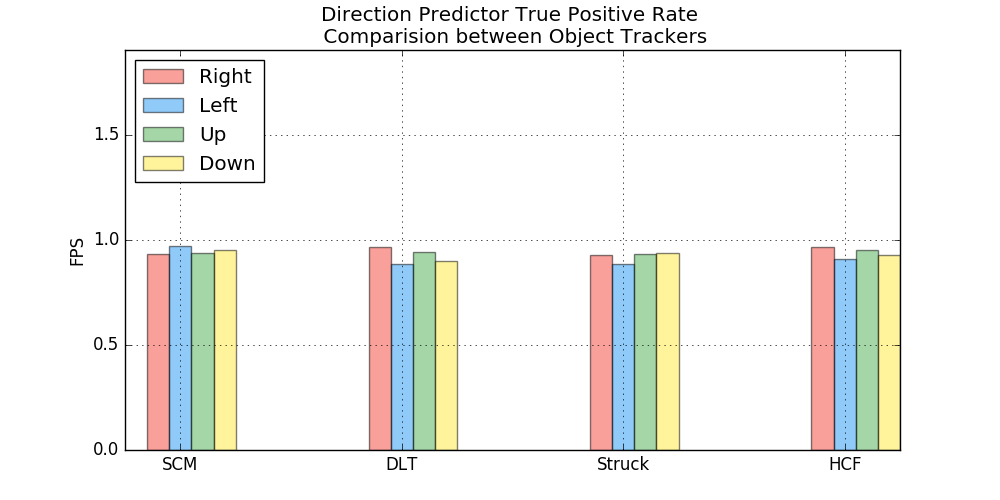} 
	\caption{TPR comparison b/w trackers with our method}
\end{figure} 

\par
\begin{figure}[h]
\centering
	\label{fig:FPR}
	\includegraphics[width=\linewidth]{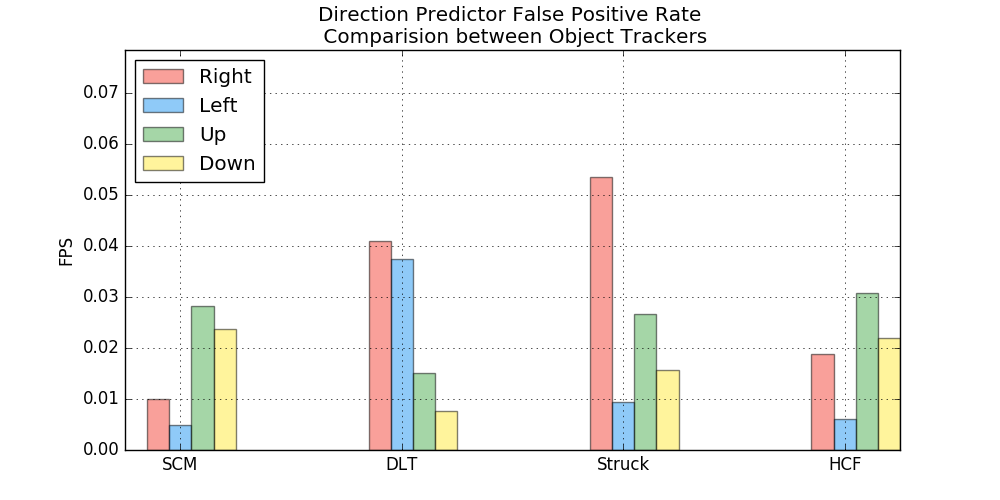}
	\caption{FPR comparison b/w trackers with our method}
\end{figure}

 Next, we have implemented our direction prediction model on each of these trackers. The \ref{fig:TPR} shows the True positive rate of the predictor algorithm. From the TPR we can infer that predictor is close to ideal direction predictor. But if we observe \ref{fig:FPR} we can see that the predictor works differently for each tracker as there movement tracking algorithms are different which leads to these differences.
\subsection{Reducing number of operations}
We have implemented the compression of CNN models on SSD object detection. For SSD architecture, we have used different classifiers for object classification, the \ref{table:comparision} shows that the even with \textbf{10 times} smaller size our model was able to able to attain the original accuracy maintaining 95\% retention.   

\begin{figure}[h]
\centering
	\label{fig:SSD output 2}
	\includegraphics[width=0.75\linewidth]{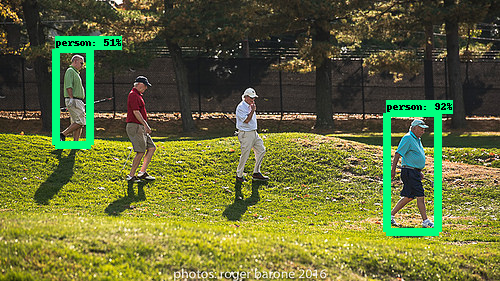}
	\caption{SSD output with partial detection of objects}
\end{figure} 

\begin{table}[h]
\setlength\tabcolsep{5pt}
  \caption{
  Comparison of mean average precision of different network architectures for SSD object detection. 
  } 
\centering 
\resizebox{\columnwidth}{!}{%
\begin{tabular}{c c c c c c} 
\hline\hline 
Object Detection  & Convolution & mAP & Million      &Model Size  &Detection Speed\\
Framework           & Model       &     & Parameters &in MB       &in FPS\\
\hline 
        & Deeplab-VGG  & 21.95\% & 33.1 & 247.5  &Out of memory \\ 
SSD 300 & Inception V2 & 22.05\% & 13.7 & 196.9  &0.153 \\ 
        & MobileNet    & 19.27\% & 6.8  & 29.2   &2.167 \\
\hline 
\end{tabular}
}
\label{table:comparision} 
\end{table}

\begin{figure}[h]
\centering
	\label{fig:SSD output 1}
	\includegraphics[width=0.75\linewidth]{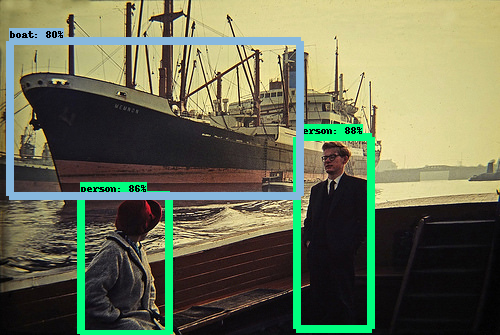}
	\caption{SSD output with complete detection of objects}
\end{figure}

\section{Conclusion and Future Work}
In this project we proposed two methods for reducing computation in visual tasks. There are other methods like network pruning which can be used to reduce the size of model without degrading the accuracy. Also for trackers we can employ parallelism to increase fps for object tracking and direction prediction. We explored only SSD architecture for object detection we can further explore other object detection frameworks like YOLO to verify our network compression postulates.

\bibliographystyle{ieeetr}
\bibliography{main.bib}
\end{document}